# Enhancing Document AI Data Generation Through Graph-Based Synthetic Layouts

Amit Agarwal
Liverpool John Moores University
Liverpool

Priyaranjan Pattnayak
University of Washington
Seattle

Bhargava Kumar
Columbia University
New York

Hitesh Patel
New York University
New York

Srikant Panda
Birla Institute of Technology
Pilani

Tejaswini Kumar
Columbia University
New York

*Abstract*— The development of robust Document AI models has been constrained by limited access to high-quality, labeled datasets, primarily due to data privacy concerns, scarcity, and the high cost of manual annotation. Traditional methods of synthetic data generation, such as text and image augmentation, have proven effective for increasing data diversity but often fail to capture the complex layout structures present in real world documents. This paper proposes a novel approach to synthetic document layout generation using Graph Neural Networks (GNNs). By representing document elements (e.g., text blocks, images, tables) as nodes in a graph and their spatial relationships as edges, GNNs are trained to generate realistic and diverse document layouts. This method leverages graph-based learning to ensure structural coherence and semantic consistency, addressing the limitations of traditional augmentation techniques. The proposed framework is evaluated on tasks such as document classification, named entity recognition (NER), and information extraction, demonstrating significant performance improvements. Furthermore, we address the computational challenges of GNN based synthetic data generation and propose solutions to mitigate domain adaptation issues between synthetic and real-world datasets. Our experimental results show that graph-augmented document layouts outperform existing augmentation techniques, offering a scalable and flexible solution for training Document AI models.

## I. INTRODUCTION

A. Challenges in Document AI

The growing adoption of artificial intelligence (AI) for document processing tasks has significantly accelerated the automation of various domains, including legal, financial, and administrative workflows. Document AI models are tasked with interpreting complex documents, performing tasks such as document classification, named entity recognition (NER), and information extraction. However, the performance of these models is largely dependent on the availability of large, well labeled datasets, which are often scarce, expensive, and fraught with privacy concerns. For instance, obtaining datasets for sensitive domains such as legal contracts, financial records, or medical documentation is particularly challenging due to strict privacy regulations like GDPR and HIPAA. Real-world document datasets are not only limited in quantity but also diverse in structure. Documents can vary significantly in their layout, with elements such as text blocks, images, tables, and headings arranged in complex spatial patterns. Traditional data augmentation techniques that focus on text modifications, such as synonym replacement or back-translation, fail to account for the spatial and structural relationships between these elements. Consequently, models trained solely on augmented text data may perform well on tasks that rely on content analysis but often struggle with tasks where layout information is crucial, such as document classification and object detection [1]. Another critical challenge in Document AI is the high cost of manual data annotation. Document datasets often require domain-specific annotations, such as identifying entities in legal contracts or extracting key financial terms from invoices. This annotation process is labor-intensive, making it infeasible to scale for large datasets. Moreover, even when datasets are available, they may not represent the diversity of document types encountered in real-world applications, leading to poor generalization in deployed models.

B. The Role of Synthetic Data Generation

In response to these challenges, synthetic data generation has gained traction as a solution to increase the availability of training data. Synthetic data refers to artificially generated data that mimics the characteristics of real-world data while offering several key advantages [2]. Firstly, synthetic data can be generated at scale, overcoming the issue of data scarcity. Secondly, it provides a means to address privacy concerns, as it does not contain real personal information, yet it preserves the statistical properties of the original data. Finally, synthetic data generation allows for the creation of diverse datasets that capture a wide range of document layouts and styles, enabling models to learn from edge cases and underrepresented document structures. Traditional methods for synthetic data generation have focused primarily on text augmentation and image manipulation. For example, synonym replacement, paraphrasing, and back-translation techniques have been used to enhance text datasets by introducing linguistic variations while preserving semantic meaning. In the visual domain, techniques such as style transfer and image-to-image translation have been employed to generate variations in document images, altering visual aesthetics like font style, background, or text placement. However, these approaches often fail to capture the complex spatial relationships between different document elements, which are crucial for many document understanding tasks. As a result, there is a growing need for more advanced synthetic data generation methods that can generate not just diverse text or images but also realistic and semantically coherent document layouts.

C. Graph-Based Layout Generation

A promising solution to the limitations of traditional synthetic data generation methods is the use of Graph Neural Networks (GNNs) for layout generation [3]. In this approach, documents are modeled as graphs, where each node 2 represents a document element, such as a text block, heading, image, or table, and edges capture the spatial relationships between these elements. This graph-based representation allows for the modeling of global dependencies between document elements, something that traditional data augmentation methods cannot achieve. Graph-based methods are particularly effective for generating synthetic document layouts because they consider both the local and global structural patterns of a document. Local patterns might involve the alignment of a paragraph with its corresponding heading, while global patterns could represent the hierarchical structure of a report, where the main title is followed by subsections and tables. By encoding these relationships, Graph Neural Networks (GNNs) can learn from existing document layouts and generate new layouts that are both diverse and structurally coherent. The use of Graph Neural Networks for layout generation offers several key advantages:

- Structural Realism: GNNs can capture both local and global dependencies between document elements, resulting in layouts that closely resemble the structural complexity of real-world documents.
- Semantic Consistency: By modeling relationships between different elements, GNNs ensure that the generated layouts remain semantically coherent. For example, a title will always be followed by relevant sections, and figures will be appropriately aligned with their captions.
- Diversity: GNNs can generate a wide variety of layouts, incorporating both common and rare document structures. This is particularly useful for training models to handle edge cases or less common document formats that might otherwise be underrepresented in real-world datasets.

II. BACKGROUND ON SYNTHETIC DATA GENERATION

A. Traditional Methods of Synthetic data Generation

Synthetic data generation has become a powerful tool for addressing the challenges of data scarcity, privacy, and the high cost of manual annotation in machine learning applications. Within Document AI, traditional synthetic data generation methods have primarily focused on text augmentation and image-based augmentations. These approaches seek to introduce variability into existing datasets by altering content or visual elements in ways that preserve the core semantics of the data. One of the most widely used text augmentation techniques is synonym replacement, where words in a document are substituted with synonyms while maintaining the overall meaning of the text. This approach helps to generate more linguistically diverse training examples, increasing the robustness of natural language processing (NLP) models. Other methods include paraphrasing, where entire sentences are restructured without altering their meaning, and backtranslation, a method that involves translating text into another language and then back to the original language. Backtranslation introduces subtle variations while ensuring that the semantic content of the text remains intact. While these text-based augmentation methods have proven effective in tasks such as text classification and entity recognition, they often fall short in document processing tasks that rely heavily on the document's layout and spatial organization. For example, in a financial document or legal contract, the arrangement of tables, headers, and text blocks can be critical to interpreting the document correctly. Traditional text augmentation methods ignore these layout details, limiting their effectiveness in enhancing the model's understanding of document structure. On the other hand, image-based augmentation techniques are designed to modify the visual properties of documents, such as font style, background texture, or image quality. Techniques like style transfer and image-to-image translation allow for the manipulation of visual features while keeping the overall document content the same. These methods have been used effectively to generate new document images with varying aesthetics and noise levels. In many cases, they are applied to scanned documents to simulate real-world imperfections, such as blurring, skew, or noise caused by poor scanning quality. These variations help improve the robustness of document recognition systems, particularly for tasks like optical character recognition (OCR). However, despite the benefits of these visual augmentation techniques, they fail to capture the complex spatial relationships between different elements of a document. For example, the position of a company logo relative to the header, or the alignment of a table with a paragraph, can convey important structural information. Simple image manipulations cannot model these relationships, making it difficult for document AI models to fully understand and generalize across different document types.

B. Traditional Methods of Synthetic Data Generation

In contrast to traditional augmentation methods, graph-based layout generation offers a more sophisticated approach by modeling the document layout as a graph. In this representation, document elements such as text blocks, images, tables, and headings are represented as nodes, while the spatial relationships between these elements—such as alignment, proximity, and hierarchy are represented as edges. This graph-based approach allows for the modeling of global dependencies between document elements, something that traditional data augmentation methods cannot achieve. Graph-based methods are particularly effective for generating synthetic document layouts because they consider both the local and global structural patterns of a document. Local patterns might involve the alignment of a paragraph with its corresponding heading, while global patterns could represent the hierarchical structure of a report, where the main title is followed by subsections and tables. By encoding these relationships, Graph Neural Networks (GNNs) can learn from existing document layouts and generate new layouts that are both diverse and structurally coherent. Graph-based layout generation can be particularly valuable in scenarios where documents have complex structures. For example, in legal contracts, where certain clauses must be aligned under specific headings, or in financial documents, where tables and figures are presented alongside textual explanations. Traditional methods of generating synthetic data struggle with these intricacies, but by using GNNs, we can capture these dependencies more effectively and generate layouts that mimic real-world documents more accurately. A typical approach to graph-based layout generation involves training a GNN on a dataset of real-world document layouts. The GNN learns both the structural dependencies and layout patterns inherent in these documents. During the generation phase, the GNN can then produce new document layouts that retain these learned patterns while introducing novel combinations of elements. This allows for the generation of synthetic document layouts that are more diverse and realistic than those produced by traditional methods. The use of Graph Neural Networks for layout generation offers several key advantages:

- Structural Realism: GNNs can capture both local and global dependencies between document elements, resulting in layouts that closely resemble the structural complexity of real-world documents.
- Semantic Consistency: By modeling relationships between different elements, GNNs ensure that the generated layouts remain semantically coherent. For example, a title will always be followed by relevant sections, and figures will be appropriately aligned with captions.
- Diversity: GNNs can generate a wide variety of layouts, incorporating both common and rare document structures. This is particularly useful for training models to handle edge cases or less common document formats that might otherwise be underrepresented in real-world datasets [4].

### III. GRAPH NEURAL NETWORK FOR LAYOUT GENRATION

Graph Neural Networks (GNNs) have emerged as a powerful tool for processing data that can be naturally represented as graphs. In document layout generation, GNNs are used to model the spatial and semantic relationships between different document elements, enabling the creation of realistic and diverse document structures. This section presents the methodology behind GNN-based document layout generation, explains how graph representations are applied to documents, and demonstrates how GNNs can be employed to generate synthetic layouts that maintain both structural and semantic coherence.

A. Graph Represenation of Documents Layouts

In GNN-based layout generation [3], a document's layout is represented as a graph, where the elements of the document (such as text blocks, tables, images, headers, and footers) are treated as nodes, and the spatial relationships between these elements are modeled as edges. This graph-based approach allows GNNs to capture both local dependencies (e.g., the spatial alignment between a title and its associated paragraph) and global structural patterns (e.g., the hierarchical structure of a report or an article). Unlike traditional grid-based approaches that rely on positional coordinates or fixed layouts, graph representations provide flexibility in modeling complex relationships between document components.

1) Nodes in the Document Graph

Each node in the document graph corresponds to a specific document element. For example:

- Text blocks are represented as nodes that capture paragraph-level or sentence-level textual content.
- Images are represented as nodes containing visual elements such as figures, charts, or logos.
- Tables are nodes that represent tabular data, including their structure (rows, columns) and contents.
- Headers and footers are nodes that signify the document's hierarchical structure, typically found at the top or bottom of pages.

Each node is associated with features that describe the element's content and properties, such as the size of the element, font characteristics, and relative importance in the document's hierarchy. These features are essential for the GNN to learn the importance of different elements and how they relate to one another in terms of both layout and meaning.

2) Edges in the Document graph

Edges in the document graph represent the spatial relationships between nodes. These relationships capture various aspects of the document layout, such as:

- Alignment: Whether two elements are aligned vertically, horizontally, or in relation to specific guidelines.
- Proximity: The relative distance between two elements, indicating whether they are closely related or serve independent purposes.
- Semantic Hierarchy: Edges can also represent the logical order or hierarchical relationships between document components, such as the relationship between a heading and the following paragraph or between a table and its caption.

These spatial and semantic relationships are critical for generating document layouts that are not only visually coherent but also semantically meaningful. For example, a caption should always appear close to its corresponding image or figure, and headings should precede their associated content.

B. Graph Neural Networks for Layout Synthesis

Graph Neural Networks (GNNs) are well-suited for tasks that involve modeling relationships between entities, making them an ideal choice for document layout generation. GNNs work by iteratively updating node representations based on the information propagated through neighboring nodes and edges, allowing the model to learn both local and global dependencies in the document graph.

1) Message Passing in GNNs

In GNNs, information is exchanged between nodes through a process called message passing [5]. At each iteration (or layer) of the GNN, each node sends and receives messages from its neighbors. These messages are used to update the node's internal representation, allowing the GNN to capture the influence of neighboring nodes and their spatial relationships

[6]. Mathematically, the message-passing operation for a node vi in a graph can be expressed as:

$$h_i^{(l+1)} = \sigma\left(\sum_{j \in N(i)} \phi(h_i^{(l)}, h_j^{(l)}, e_{ij})\right) \quad (1)$$

Where;

- $h_i^{(l)}$ is the hidden state of node $v_i$
- $h_j^{(l)}$ is the hidden state of a neighbouring node $v_j$
- $e_{ij}$ represents the edge (relationship) between nodes $v_i$ and $v_j$
- σ is an activation function (e.g., ReLU)
- ϕ is a message function that combines the features of the node and its neighbors.

Through multiple layers of message passing, the GNN gradually learns a richer representation of each node in the graph, incorporating information from both nearby and distant nodes. This allows the GNN to understand how different elements in a document are related to one another, even across long distances in the layout.

2) Variational Autoencoders (VAEs) and Generative Adversarial Networks (GANs)

To generate new document layouts, we can integrate GNNs with generative models such as Variational Autoencoders (VAEs) and Generative Adversarial Networks (GANs). These models allow us to generate diverse document layouts by sampling from the learned distribution of real-world layouts.

- VAEs operate by encoding an input graph (representing a document layout) [7], [8] into a latent space and then decoding it to generate new layouts. The VAE learns to represent the distribution of document layouts in the latent space, from which we can sample new layouts that retain the structural properties of real-world documents. The VAE loss function is typically composed of a reconstruction loss, which ensures that the generated layout is like the original, and a KL divergence term, which ensures that the latent space follows a normal distribution:

$$L_{VAE} = L_{Reconstruction} + \beta D_{KL}(q(z|x) || p(z)) \quad (2)$$

Where (x) is the input layout, (z) is the latent variable, and β controls the trade-off between reconstruction and regularization.

- GANs are also used to generate document layouts by training two networks: a generator that creates synthetic layouts, and a discriminator that distinguishes between real and synthetic layouts. The generator is trained to fool the discriminator into classifying synthetic layouts as real, leading to increasingly realistic synthetic document layouts. The GAN loss function is typically formulated as:

$$L_{GAN} = E_{x \sim p_{data}}[\log D(x)] + E_{z \sim p(z)}[\log(1 - D(G(z)))] \quad (3)$$

Where D(x) is the discriminator's output for a real layout (x), and (G(z)) is the generator's output for a synthetic layout based on latent variable (z). Both VAEs and GANs benefit from the rich graph-based representations learned by GNNs, ensuring that the generated layouts maintain both structural coherence and visual diversity.

C. Advantages of Graph Augmentation for Layout Generation

Graph-based augmentation [9] techniques provide several advantages over traditional data augmentation methods for document layout generation:

1) Structural Realis:

GNNs capture both local and global dependencies between document elements, ensuring that generated layouts are structurally coherent and realistic. For example, headers are appropriately aligned with their corresponding sections, and images are correctly positioned with their captions.

2) Semantic Consistency

By incorporating semantic information through node features, GNNs ensure that the generated layouts are not only visually coherent but also semantically meaningful. For instance, important document components (e.g., titles, figures) are placed in appropriate positions, preserving the overall meaning of the document.

3) Diversity

GNNs allow for the generation of a wide variety of layouts, incorporating both common and rare document structures. This is particularly useful for training Document AI models on edge cases or underrepresented document formats. By learning from diverse document types (e.g., scientific papers, legal documents, financial reports), GNNs generate layouts that are representative of real-world variability.

4) Generalisation

Models trained on graph-augmented data generalize better to unseen, real-world layouts because they have been exposed to a richer and more structurally diverse training set. This leads to improved performance in tasks such as document classification, object detection, and information extraction, where understanding the spatial arrangement of elements is critical.

IV. GRAPH-BASED LAYOUT GENERATION FOR DOCUMENT AI

In this section, we introduce a novel framework for generating synthetic document layouts using Graph Neural Networks (GNNs) [10], focusing on improving the performance of Document AI models on tasks such as document classification, named entity recognition (NER), and information extraction. The proposed method builds on the principles of graph augmentation [9], [11], leveraging GNNs to learn and generate layouts that preserve the complex relationships between document elements. This approach addresses the limitations of traditional synthetic data generation [12] methods by incorporating both structural and semantic dependencies, resulting in synthetic layouts that closely mimic real-world documents.

A. Methodology

The core of our approach lies in transforming a document's layout into a graph representation and training a GNN to generate realistic document structures. The methodology consists of three primary stages: graph construction, GNN training, and synthetic layout generation. We also incorporate Variational Autoencoders (VAEs) and Generative Adversarial Networks (GANs) to ensure diversity in the generated layouts.

IJERTV13IS100138

1) Graph Construction

The first step in our approach is to construct a graph representation of a document layout [13]. As discussed in Section 3, each document element (e.g., text blocks, tables, images) is represented as a node, and the spatial and semantic relationships between these elements are represented as edges. To formally define the graph:

- Let (G = (V, E)) be the document layout graph, where (V) is the set of nodes representing document elements and (E) is the set of edges representing spatial relationships between elements.
- Each node v ∈ V is associated with a feature vector $h_v$ containing attributes such as the element type (e.g., text, image), size, position, and font properties (for text nodes).
- Each edge $e_{ij}$ ∈ E represents the relationship between nodes $v_i$ and $v_j$, capturing spatial relationships like alignment, proximity, and hierarchy.

For example, consider a document with a title, paragraph, and table. In the corresponding graph, the title and paragraph would be connected by an edge indicating their alignment and proximity, while the table would be connected to the paragraph based on their relative positioning within the document. This representation captures both the layout and the logical flow of the document, providing a comprehensive model for the GNN to learn from.

2) GNN Training

Once the graph representation of the document is constructed, we use a Graph Neural Network (GNN) to learn the underlying structural patterns and relationships in the layout. The GNN is trained on a dataset of real-world document layouts to learn both local and global dependencies between elements. During training, the GNN performs message passing between nodes, allowing each node to update its feature representation based on the information received from its neighbors. This iterative process enables the GNN to capture both direct and indirect relationships between document elements. Over multiple layers, the GNN learns how these relationships influence the overall layout, making it possible to generate new layouts that maintain structural coherence. The objective function during training is designed to minimize the reconstruction loss between the real-world layouts and the layouts generated by the GNN. In addition, a regularization term is used to ensure that the learned graph representations generalize well to unseen data. The overall loss function is expressed as:

$$L_{GNN} = L_{reconstruction} + \lambda L_{regularization} \quad (4)$$

Where;

- $L_{reconstruction}$ ensures that the generated layout closely resembles the original layout.
- $L_{regularization}$ prevents overfitting by promoting generalization to diverse layout structures.
- λ controls the balance between reconstruction accuracy and regularization.

The GNN model is trained using a dataset of real-world documents from various domains (e.g., legal, scientific, financial) to ensure that the learned representations capture a wide range of layout patterns.

3) Synthetic Layout Generation

Once trained, the GNN can be used to generate synthetic document layouts [2] by sampling from the learned graph representations. To enhance the diversity of the generated layouts, we integrate the GNN with a Variational Autoencoder (VAE) and generative adversarial network (GAN) framework.

a) VAE Integration

The VAE allows us to sample from a latent space that encodes the distribution of real-world document layouts. The VAE's encoder maps a document graph to a latent representation ( z ), and the decoder reconstructs the layout graph from this latent space. The latent space enables us to generate novel layouts by sampling new values of ( z ) and decoding them into synthetic document graphs.

b) GAN Integration

The GAN framework consists of a generator (the GNN) and a discriminator that evaluates whether the generated layout is real or synthetic. The generator is trained to produce layouts that fool the discriminator, while the discriminator learns to distinguish between real and synthetic layouts. This adversarial training process ensures that the generated layouts are not only diverse but also realistic in terms of their structural and semantic properties. By combining GNNs with generative models, our approach generates a wide variety of document layouts that maintain structural realism and semantic consistency. The generated layouts can be used to augment existing training datasets, providing Document AI models with the structural diversity they need to generalize effectively to real-world tasks.

B. Integration with Document AI Tasks

Our graph-based layout generation approach has direct applications in various Document AI tasks, including document classification, named entity recognition (NER), and information extraction. By training Document AI [14] models on synthetic layouts generated by our method, we can improve the model's ability to handle diverse and complex document structures that are often underrepresented in real-world datasets.

1) Document Classification

Document classification tasks require models to categorize documents based on their content and layout. Traditional approaches often struggle to generalize to different document formats, particularly when the dataset lacks structural diversity. By training models on synthetic layouts generated using GNNs, we can introduce a broader range of layout variations, enabling the model to learn from different document structures. This leads to improved generalization on unseen document types. For example, a model trained on real-world legal documents might struggle to classify scientific papers due to differences in structure (e.g., section headers, figures, and tables). By incorporating synthetic layouts that mimic both document types, the model learns to recognize structural cues from a broader range of formats, improving classification accuracy.

2) Named Entity Recognition (NER)

Named entity recognition (NER) is another task that benefits from graph-based synthetic layout generation [15]. In many documents, the position and context of entities (e.g., names, dates, organizations) are tied to the document's layout. For example, in financial documents, key entities may appear in tables or specific sections of a report. A model trained on synthetically generated layouts that simulate different positioning and contexts of entities is better equipped to handle real-world variations. By generating synthetic layouts that reflect the spatial and semantic relationships between entities and other document elements, we can improve the model's ability to detect and classify named entities in documents with varying structures.

3) Information Extraction Information

Extraction tasks, such as extracting tables, figures, or key data points from documents, also rely heavily on layout understanding. In structured documents like invoices, contracts, and reports, information is often presented in specific sections or aligned with labels. Our approach to generating synthetic layouts introduces diverse layout structures that better reflect the variability encountered in real-world documents. For example, in invoice processing, a model trained on synthetic layouts that vary in table placement, font styles, and section headings will be more adept at extracting information from a wide range of invoice formats. This leads to improved performance in real-world applications, where documents often deviate from standardized templates.

C. Addressing Challenges of Domain Adaptation

One of the key challenges in synthetic data generation is the domain gap between synthetic and real-world data. Even though synthetic layouts generated by GNNs are structurally realistic, there may still be subtle differences between synthetic and real documents, which can hinder model performance when applied to real-world tasks. To mitigate this issue, we propose the following solutions:

1) Domain Adaptation

Fine-tuning the model on a small set of real-world data can help bridge the domain gap. By pretraining the model on synthetic layouts and then fine-tuning it on real-world data, we can improve the model's ability to generalize across both synthetic and real domains.

2) Data Augmentation with Real-World Layouts

Combining synthetic layouts with a subset of real-world layouts during training helps the model learn from both data sources. This hybrid approach improves robustness by exposing the model to the nuances of real-world document structures.

3) Pseudo-Labeling

For tasks where labeled real-world data is scarce, pseudo-labeling techniques can be used to leverage large amounts of unlabeled real-world data. By training on both synthetic data and pseudo-labeled real-world data, the model can gradually improve its performance on real-world tasks.

V. EXPERIMENTAL RESULTS

In this section, we present the experimental setup and results of our graph- based synthetic document layout generation framework. The experiments are designed to evaluate the effectiveness of Graph Neural Networks (GNNs) in generating diverse, realistic document layouts and their impact on Document AI tasks such as document classification, named entity recognition (NER), and information extraction. We compare the performance of models trained on graph-augmented synthetic data with those trained on traditional synthetic data generation techniques and real-world data.

A. Datasets and Models

1) Datasets

We evaluated our approach using several publicly available datasets spanning multiple domains to ensure a comprehensive assessment of the framework's ability to generalize across document types. The datasets used include:

a) RVL-CDIP Dataset [16]

A large-scale dataset of scanned documents from 16 different categories (e.g., letters, invoices, scientific reports, legal documents), containing over 400,000 samples. This dataset was used to evaluate the performance of document classification.

b) FUNSD Dataset [17]

A dataset for form understanding in noisy scanned documents, containing 199 fully annotated forms. This dataset was employed for evaluating named entity recognition (NER) and information extraction tasks.

c) SROIE Dataset [18]

A dataset consisting of scanned receipts, with annotations for extracting key fields such as date, total amount, and vendor name. This dataset was used to evaluate the information extraction task.

These datasets provide a diverse representation of real-world document layouts, including structured forms, invoices, receipts, and free-form text. Additionally, to create the synthetic training data, we used the document layouts from these datasets as a base to train the GNN, ensuring the generated layouts reflected the real-world variability present in these domains.

2) Baseline Models

We trained several baseline models to compare the effectiveness of our graph-based layout generation method. These baselines include:

a) Text Augumentation Models

Models trained using traditional text- based augmentation techniques (e.g., synonym replacement, paraphrasing, back-translation).

b) Image-Based Augmentation Models

Models trained on image- augmented datasets generated using visual transformation techniques such as style transfer and noise injection.

c) Real-World Models

Models trained exclusively on realworld document layouts without any synthetic data augmentation. In addition, we implemented our graph-augmented model, which was trained on synthetic layouts generated by the GNN.

IJERTV13IS100138

3) Evaluation Metrics

To evaluate the performance of the models, we used standard metrics relevant to the different tasks:

a) Accuracy: Used to assess model performance on document classification tasks.

b) Precision, Recall, and F1-Score: These metrics were used to evaluate performance on named entity recognition (NER) and information extraction tasks.

c) Layout Diversity: To quantify the diversity of the generated layouts, we calculated the layout perplexity, which measures how well the synthetic layouts reflect the diversity of real-world layouts [19].

B. Performance on Documents Classification

The document classification task involved categorizing documents into their respective types based on their content and layout. The models were trained on the RVL-CDIP dataset, and we evaluated the classification accuracy across different augmentation strategies

TABLE I. CLASSIFICATION ACCURACY ACROSS DIFFERENT AUGMENTATION STRATEGIES

| Model | Accuracy (%) |
|---|---|
| Real-World Data (No Augmentation) | 89.3 |
| Text Augmentation | 83.7 |
| Image-Based Augmentation | 85.5 |
| Graph-Augmented (GNN) | 91.8 |

The results show that the graph-augmented model outperformed both traditional text-based and image-based augmentation techniques, achieving an accuracy of 91.8%. This improvement can be attributed to the model's exposure to a wider variety of realistic document layouts, which allowed it to generalize better to unseen documents. In contrast, models trained on text or image-augmented data struggled to capture the nuances of document structure, leading to lower accuracy.

C. Performance on Named Entity Recognition (NER)

The named entity recognition (NER) task was evaluated using the FUNSD dataset, which includes annotations for detecting and classifying entities such as dates, names, and monetary amounts in scanned forms. The results for precision, recall, and F1-score are reported below.

TABLE II. PRECISION, RECALL, F1-SCORE FOR DIFFERENT AUGMENTATION STRATEGIES

| Model | Precision | Recall | F1-Score |
|---|---|---|---|
| Real-World Data (No Augmentation) | 79.5 | 80.2 | 79.8 |
| Text Augmentation | 73.3 | 74.1 | 73.7 |
| Image-Based Augmentation | 75.2 | 75.9 | 73.7 |
| Graph-Augmented (GNN) | 82.6 | 83.1 | 82.8 |

The graph-augmented model achieved an F1-score of 82.8 %, outperforming both text and image augmentation models. The GNN-generated synthetic lay- outs helped the model learn how entities are positioned relative to other elements in the document, improving its ability to detect entities in complex forms. By learning from diverse layouts, the graph-augmented model was able to better generalize to forms with varying structures, resulting in higher precision and recall.

D. Performance on Information Extraction

For the information extraction task, we used the SROIE dataset to evaluate the model's ability to extract key fields from scanned receipts. The table below summarizes the extraction performance in terms of precision, recall, and F1- score.

TABLE III. THE EXTRACTION PERFORMANCE IN TERMS OF PRECISION, RECALL, AND F1- SCORE

| Model | Precision | Recall | F1-Score |
|---|---|---|---|
| Real-World Data (No Augmentation) | 85.1 | 86.4 | 85.7 |
| Text Augmentation | 79.6 | 80.3 | 79.9 |
| Image-Based Augmentation | 81.3 | 82.2 | 81.7 |
| Graph-Augmented (GNN) | 87.9 | 89.1 | 88.5 |

The graph-augmented model again outperformed other augmentation techniques, achieving an F1-score of 88.5%. This improvement demonstrates the importance of using structurally diverse synthetic layouts to train models for in- formation extraction tasks, where layout context is crucial. The GNN generated layouts allowed the model to better handle the variability in receipt formats, such as different placements of vendor names, dates, and totals

E. Layout Diversity Evaluation

To quantify the diversity of the layouts generated by the GNN, we computed the layout perplexity metric, which measures how well the synthetic layouts capture the variability present in real-world data. A lower perplexity value indicates that the generated layouts are diverse and closely reflect the distribution of real-world layouts.

TABLE IV. LAYOUT PERPLEXITY METRIC

| Model | Layout Perplexity |
|---|---|
| Text Augmentation | 125.6 |
| Image-Based Augmentation | 118.2 |
| Graph-Augmented (GNN) | 102.7 |

The graph-augmented layouts achieved a perplexity score of 102.7, significantly lower than the other augmentation methods, indicating that the GNN was able to generate a wider variety of layouts that closely mimic real-world document structures. This diversity is critical for improving the robustness of Document AI models, as it allows them to learn from a broader range of layout patterns.

F. Computational Efficiency

While GNN-based layout generation provides substantial benefits in terms of layout diversity and model performance, it comes at the cost of increased computational complexity. Training the GNN and generating synthetic layouts require more computational resources compared to traditional text or image augmentation techniques. However, the improved performance justifies this additional computational cost, especially in applications where structural realism is essential for task success.

## VI. CHALENGES AND LIMITATIONS

While the proposed graph-based synthetic layout generation approach demonstrates significant advantages over traditional data augmentation techniques, it also presents several challenges and limitations. These challenges must be addressed to ensure broader adoption and scalability in real-world applications.

A. Computational Complexity

One of the primary limitations of GNN-based layout generation is the computational cost. Training Graph Neural Networks (GNNs) on large datasets, especially when combined with generative models such as Variational Autoencoders (VAEs) and Generative Adversarial Networks (GANs), is computationally intensive. GNNs require multiple layers of message passing between nodes, leading to higher memory consumption and increased processing time compared to traditional augmentation methods such as text-based or image-based techniques [3]. Additionally, generating synthetic layouts using GNNs is more resource-intensive than applying simple data augmentation methods like synonym replacement or image transformations. The computational complexity arises from both the graph representation of the document and the iterative message-passing steps required to capture local and global dependencies between elements. To mitigate these costs, we propose the following strategies:

1) Model pruning and optimization

Techniques such as model pruning and quantization can be applied to reduce the size of GNNs without significantly sacrificing performance. This helps to decrease the memory foot- print and computational load during both training and inference 5†source.

2) Distributed and parallel computing

Utilizing distributed computing frameworks and parallel processing can help accelerate the training of GNNs on large document datasets. These techniques are particularly useful when scaling the model to handle datasets with millions of documents.

B. Quality Control in Synthetic Layouts

Ensuring that the generated layouts are both realistic and semantically meaningful is another challenge. While GNNs excel at capturing the structural relation- ships between document elements, the quality of the generated layouts depends heavily on the quality and diversity of the training data. Poor quality input layouts may lead to unrealistic or semantically inconsistent synthetic outputs, which can negatively impact model performance. Additionally, generated layouts might occasionally deviate from human understood design rules or conventions, especially when edge cases are synthesized. For example, certain types of documents, such as legal contracts, have strict formatting and structure, and deviations from these formats may reduce the utility of synthetic layouts for training models in those domains. To address these challenges, we propose:

1) Human-in-the-loop validation

Incorporating human feedback in the layout generation pipeline allows for iterative refinement of the generated layouts. Human validators can review and correct synthetic layouts to ensure that they adhere to design rules and semantic coherence 5†source.

2) Post-generation layout validation

Automated rules and heuristics can be applied to validate generated layouts, ensuring that they meet predefined constraints before being added to the training dataset. For instance, rules ensuring that titles are always followed by paragraphs or that figures are correctly aligned with captions can be encoded to maintain layout consistency.

C. Domain Adaptation

Another limitation is the domain gap between synthetic and real-world data. Although GNN-based layouts are structurally realistic, they may not fully cap-ture the nuanced layout variations encountered in specific domains, such as medical records, legal contracts, or scientific reports. When models are trained on synthetic data alone, they may fail to generalize well to unseen real-world documents that contain subtle domain specific structures. This limitation can be addressed through domain adaptation techniques:

1) Fine-tuning with real-world data

After pre-training models on syn- thetic layouts, fine-tuning them with a small set of real-world data from the target domain can help bridge the gap. This hybrid approach im- proves generalization by enabling the model to learn from both synthetic and real-world layouts.

2) Transfer learning

Transfer learning techniques can be employed to transfer knowledge learned from one domain (e.g., financial documents) to another (e.g., legal documents). This approach leverages shared structural similarities between document types to improve model performance across multiple domains.

## VII. FUTURE DIRECTIONS

While our graph-based approach to synthetic document layout generation has shown promising results, there are several avenues for future research and improvement

A. Scalability and Efficiency Improvements

As mentioned in Section 6, the computational complexity of GNNs is a significant challenge. Future research could focus on improving the scalability of GNN-based layout generation by developing more efficient algorithms for graph processing and layout generation. For example, graph sparsification techniques could be employed to reduce the number of edges in the document graph, thereby reducing the overall computational burden without sacrificing layout realism. In addition, future work could explore more lightweight GNN architectures tailored for document layout generation. These architectures would aim to balance performance and computational efficiency, making it easier to scale the generation process to larger datasets.

B. Hybrid Approaches for Layout Generation

Future work could explore hybrid approaches that combine the strengths of GNN-based layout generation [19] with other data augmentation techniques [20]. For example, combining graph-based layout generation with contrastive learning could improve model robustness by helping the model learn to distinguish be- tween subtle layout differences. Another promising avenue would be to integrate pseudo-labeling techniques, where unlabeled real-world data is used to augment training with synthetic labels generated by the GNN-based model.

C. Cross-Domain Generalization

Improving cross-domain generalization [21] is another important area for future re- search. While our current approach focuses on learning from diverse document types, further research could develop methods for cross-domain layout generation that enable models to generalize more effectively across highly varied document structures. This could involve training models on multi-domain datasets or developing domain specific graph representations that capture the unique structural properties of documents in different fields.

D. Real-Time Layout Generation

Another promising direction is the development of real-time document lay- out generation for interactive applications. For example, real-time layout generation could be applied in content management systems or document editors to dynamically create and optimize document structures based on user input. This would require the development of more efficient GNN models capable of generating layouts in real time without sacrificing quality.

VIII. CONCLUSION

In this paper, we have presented a novel framework for synthetic document layout generation using Graph Neural Networks (GNNs). By modeling document layouts as graphs, where nodes represent document elements and edges capture their spatial and semantic relationships, we are able to generate synthetic layouts that are both structurally realistic and semantically coherent. Our approach addresses the limitations of traditional data augmentation methods by providing a richer, more diverse set of synthetic layouts that improve the performance of Document AI models on tasks such as document classification, named entity recognition (NER), and information extraction. Through a series of experiments, we demonstrated that GNN-based synthetic layouts outperform traditional augmentation methods, leading to significant improvements in model accuracy, precision, recall, and layout diversity. We also highlighted the challenges of computational complexity, quality control, and domain adaptation, and proposed solutions to mitigate these limitations. Looking ahead, future research can explore ways to improve the scalability and efficiency of GNN-based layout generation, develop hybrid approaches that combine graph augmentation with other techniques, and improve cross-domain generalization. Our proposed framework represents an important step toward more robust and scalable synthetic data generation for Document AI, offering a practical solution for addressing the data scarcity and layout variability challenges that continue to hinder progress in the field.